\documentclass[lettersize,journal]{IEEEtran}

\usepackage{amsmath,amsfonts,amssymb}
\usepackage{mathrsfs}      
\usepackage{bm}

\usepackage{algorithmic}
\usepackage{algorithm}
\usepackage{color} 
\usepackage{orcidlink}
\usepackage{array}
\usepackage{booktabs}
\usepackage{multirow}
\usepackage{tabularx}

\usepackage{graphicx}
\usepackage[caption=false,font=normalsize,labelfont=sf,textfont=sf]{subfig}

\usepackage{textcomp}
\usepackage{stfloats}
\usepackage{url}
\usepackage{pifont}        
\usepackage{cite}

\newcolumntype{C}{>{\centering\arraybackslash}X}

\newcommand{\eg}{\textit{e.g.}}
\newcommand{\ie}{\textit{i.e.}}

\begin{document}

\title{Towards Domain-Generalized Open-Vocabulary Object Detection: A Progressive Domain-invariant Cross-modal Alignment Method}

\author{Xiaoran~Xu~\orcidlink{0009-0001-8350-2828},
        Xiaoshan~Yang~\orcidlink{0000-0001-5453-9755},
        Jiangang~Yang~\orcidlink{0000-0002-0464-8336},
        Yifan~Xu~\orcidlink{0000-0003-2467-888X},
        Xudong~Yao~\orcidlink{0000-0002-0894-0705},
        Jian~Liu~\orcidlink{0000-0001-8338-8749},
        and~Changsheng~Xu~\orcidlink{0000-0001-8343-9665}, ~\IEEEmembership{Fellow,~IEEE}
\thanks{X. Xu, X. Yang, Y. Xu, and C. Xu are with the MAIS, Institute of Automation, Chinese Academy of Sciences, Beijing, China, and also with the School of Advanced Interdisciplinary Sciences, University of Chinese Academy of Sciences, Beijing, China (e-mail: xuxiaoran22@mails.ucas.ac.cn; \{yifan.xu, xiaoshan.yang, changsheng.xu\}@nlpr.ia.ac.cn).}
\thanks{J. Yang and J. Liu are with the Institute of Microelectronics, University of Chinese Academy of Sciences, Beijing, China (e-mail: \{yangjiangang, liujian\}@ime.ac.cn).}%
\thanks{Xudong Yao is with Tianjin University of Technology, Tianjin, China
(e-mail: yaoxudong95@outlook.com).}
\thanks{Corresponding author: Xiaoshan Yang.}}



\maketitle

\begin{abstract}
Open-Vocabulary Object Detection (OVOD) has achieved remarkable success in generalizing to novel categories. However, this success often rests on the implicit assumption of domain stationarity. In this work, we revisit the OVOD paradigm and study a key vulnerability: the fragile coupling between visual manifolds and textual embeddings under distribution shifts. We first formulate Domain-Generalized Open-Vocabulary Object Detection (DG-OVOD) as an evaluation protocol for open-vocabulary recognition under visual shifts. Through empirical analysis, we observe that visual shifts can destabilize the latent cross-modal space, causing novel-category visual signals to drift away from their semantic anchors. Motivated by these observations, we propose Progressive Domain-invariant Cross-modal Alignment (PICA). PICA departs from uniform training by introducing a multi-level curriculum based on ambiguity and signal strength. It constructs a quality-adjusted curriculum over pseudo-word prototypes, refined by sample reliability and visual consistency, to encourage more stable cross-domain modality alignment. Our findings suggest that OVOD robustness under domain shifts is closely linked to the stability of the latent cross-modal alignment space. Our work provides a DG-OVOD evaluation protocol and a practical perspective on building more generalizable open-vocabulary systems beyond static laboratory conditions.
\end{abstract}


\section{Introduction}
\label{sec:intro}
\IEEEPARstart{W}{ith} the rapid advancement of Vision-Language Models (VLMs)~\cite{radford2021learning,jia2021scaling,kim2021vilt,chen2020uniter,wang2021simvlm}, Open-Vocabulary Object Detection (OVOD) has emerged as a powerful paradigm that transcends traditional closed-set detection~\cite{gu2021open, zareian2021open, zhou2022detecting, zhong2022regionclip, cheng2024yoloworld, minderer2022simple}. By projecting natural language category descriptions into a continuous semantic embedding space and aligning them with visual features~\cite{radford2021learning,minderer2022simple,liu2024grounding, cheng2024yoloworld}, OVOD enables recognition of unseen categories at inference time. This cross-modal alignment mechanism has become the foundational principle behind modern open-vocabulary detectors~\cite{zareian2021open}. This paradigm not only substantially broadens the scope of object detection but also underpins key real-world applications such as autonomous driving and open-world robotic perception.

\begin{figure*}[!t]
  \centering
  \subfloat[]{\includegraphics[width=0.49\linewidth]{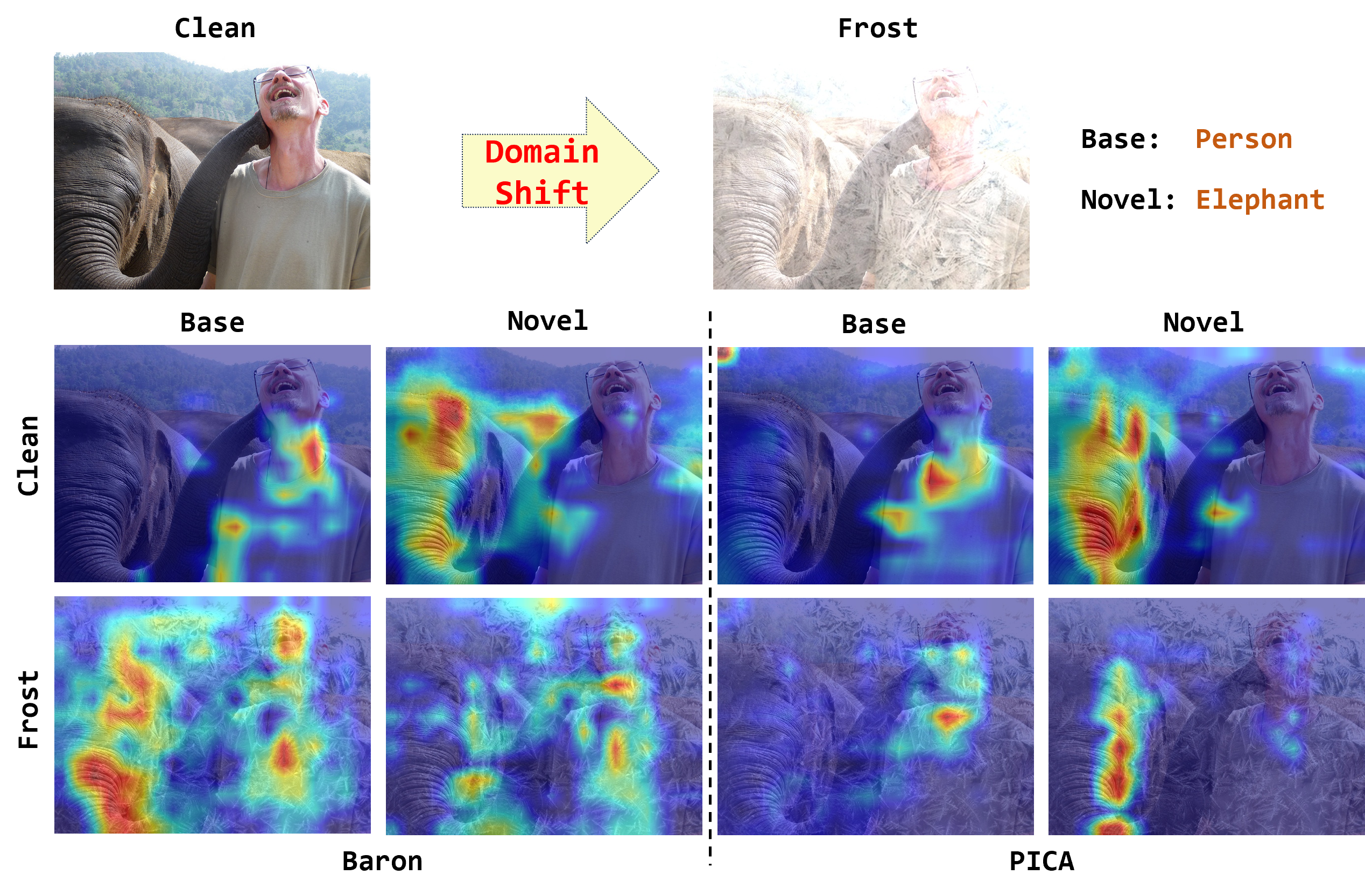}%
  \label{fig:heatmap}}
  \hfil
  \subfloat[]{\includegraphics[width=0.49\linewidth]{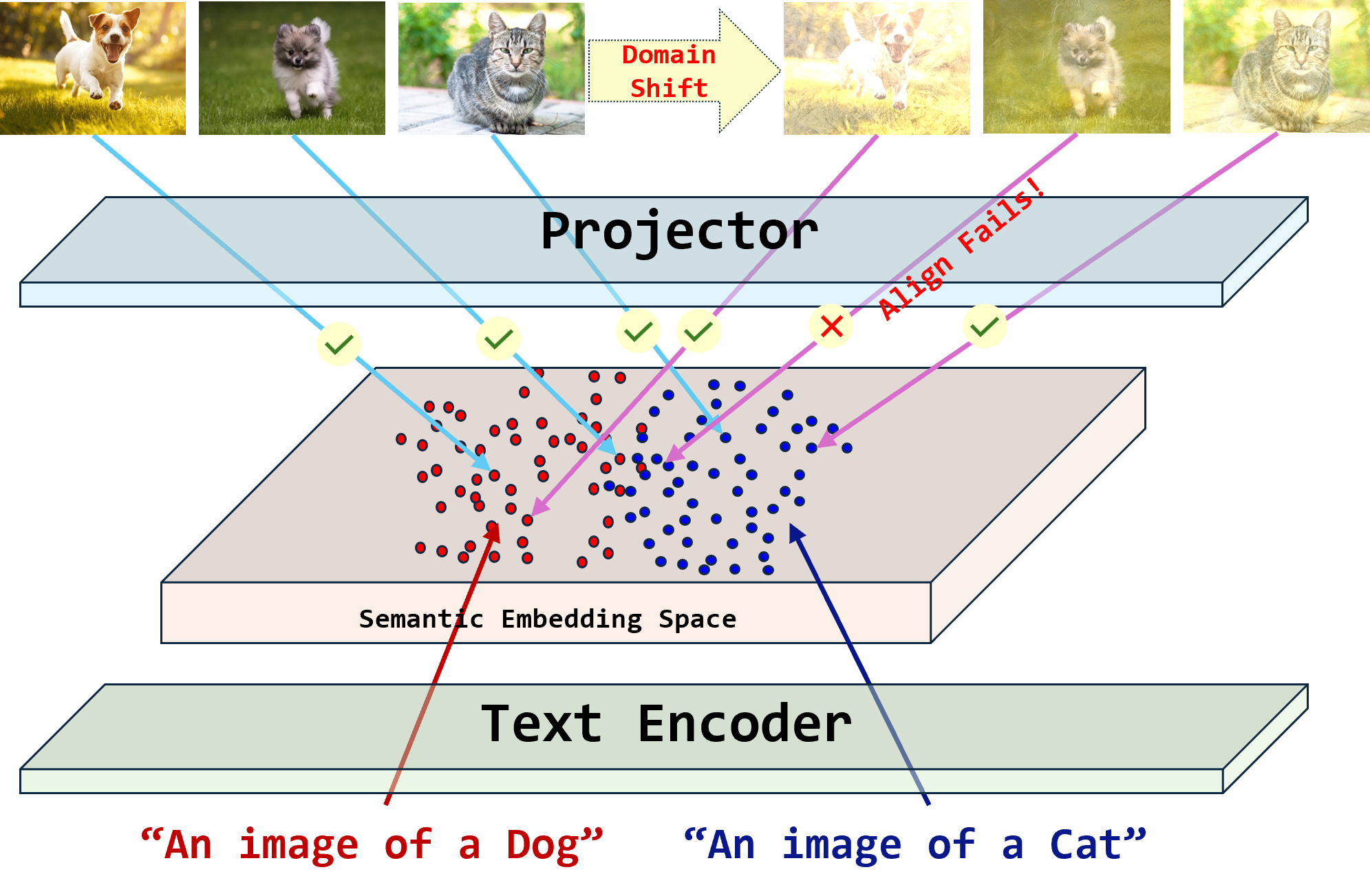}%
  \label{fig:short-b}}
  \caption{Cross-modal alignment analysis. 
  (a) Visualization of model attention under clean and frost conditions. \textit{Person} is a base category; \textit{Elephant} is a novel category. PICA preserves more stable and object-centric attention patterns across both base and novel categories compared to Baron~\cite{wu2023aligning}. (b) Cross-modal alignment degradation under visual shift is highly sample-dependent: the difficulty of preserving cross-modal alignment under domain shifts varies across different samples.}
  \label{fig:compare}
\end{figure*}

Despite the success of continuous semantic embeddings in novel-category recognition~\cite{radford2021learning,minderer2022simple,liu2024grounding, cheng2024yoloworld}, real-world environments such as autonomous driving under fog or rain can substantially alter visual appearance. Such shifts may weaken modality alignment and lead to performance degradation~\cite{shangguan2024crossdomain,chhipa2024open,singha2024unknown,zhu2024vision,chen2023benchmarking}.
In essence, while base categories may rely on robust visual priors learned during training, novel-category detection depends strongly on the integrity of the shared latent space at inference time~\cite{zareian2021open,gu2021open}. Therefore, we argue that the practical reliability of OVOD is closely related to the stability of cross-modal alignment under domain shifts.
Although this stability is important for OVOD generalization, it remains insufficiently examined, partly because conventional OVOD methods have not been systematically evaluated under diverse domain shifts.

To empirically examine this vulnerability, we analyze the projection-based alignment paradigm~\cite{zareian2021open, minderer2022simple, wu2023aligning, wu2023cora, wang2024sia, wang2025ov, bangalath2022bridging}, a widely adopted OVOD approach due to its simplicity and strong zero-shot transfer.
Despite its effectiveness, the projection is learned from source-domain statistics and is therefore susceptible to distribution shifts that can destabilize cross-modal alignment.
Taking Baron~\cite{wu2023aligning} as a representative example, Fig.~\ref{fig:heatmap} shows that domain shifts degrade spatial grounding for both base and novel categories. For the base class (Person), attention under frost becomes more dispersed but still retains coarse object-centric localization. In contrast, the novel class (Elephant) exhibits stronger distortion: activations detach from the object boundary and shift toward background textures and frost artifacts.
This disparity suggests that although spatial grounding is generally perturbed, the degradation is amplified for novel categories, whose predictions depend more heavily on stable cross-modal alignment than on learned visual priors~\cite{zareian2021open,gu2021open,zhou2022detecting,pham2024lp,wu2023aligning,minderer2022simple}. This observation highlights a vulnerability in current OVOD: the mechanism that enables open-vocabulary expansion can also become a source of failure under distribution shifts if cross-modal correspondence is not stable.

To address this vulnerability, this paper introduces Domain-Generalized Open-Vocabulary Object Detection (DG-OVOD), a unified evaluation protocol for assessing open-vocabulary generalization under visual distribution shifts. This task is non-trivial because simply integrating existing domain adaptation or domain generalization methods into OVOD does not directly address sample-specific cross-modal reliability.
As shown in Fig.~\ref{fig:short-b}, some regions suffer from signal collapse, while others face boundary confusion, where the margin to incorrect categories shrinks. This heterogeneity means that uniform training treats reliable and unreliable regions indiscriminately, which can destabilize cross-modal alignment where novel categories are most vulnerable.
Collectively, these issues suggest that improving alignment stability is important for maintaining reliable implicit supervision for novel categories under domain shift.

To mitigate sample-specific distribution shifts and alleviate the degradation of cross-modal alignment in OVOD, we introduce Progressive Domain-invariant Cross-modal Alignment (PICA).
PICA evaluates each region along two complementary dimensions that characterize these failure modes:
(1) ambiguity proxy $h_{r_i}$, which estimates boundary confusion by capturing how easily the region can be confused with hard negatives, and
(2) signal strength proxy $q_{r_i}$, which estimates signal collapse by quantifying the confidence of its positive cross-modal alignment.
These two proxies are combined into a quality-adjusted difficulty score that determines each region's training priority. During training, PICA prioritizes samples with low ambiguity and sufficient alignment signal in the early stages, and progressively incorporates harder or less reliable instances as the projection layer becomes better calibrated.
This progressive stratification is designed to reduce the influence of uncertain image-text associations in early training, while preserving informative yet challenging samples for later refinement.
As qualitatively illustrated in Fig.~\ref{fig:heatmap}, PICA mitigates shift-induced spatial grounding degradation, with a particularly clear improvement for the novel category \textit{Elephant}.

In summary, our work makes the following contributions:

\begin{itemize}
    \item We formalize Domain-Generalized Open-Vocabulary Object Detection (DG-OVOD), a systematic unified evaluation protocol designed to quantify the impact of distribution shifts on open-vocabulary generalization.
    \item Motivated by our findings, we propose PICA, a training-only progressive alignment strategy that stabilizes pseudo-word prototypes by jointly considering signal strength and boundary ambiguity.
    \item We establish a DG-OVOD evaluation suite by combining synthetic corruptions with an OV-COCO-aligned split of the existing COCO-O benchmark, and systematically evaluate representative OVOD methods under both synthetic and natural OOD domains.
\end{itemize}

\section{Related Work}
\label{sec:relatedwork}
\subsection{Open-Vocabulary Object Detection}
Open-vocabulary object detection (OVOD) enables detectors trained on a closed set of base categories to recognize novel classes at inference. Since annotations for novel categories are unavailable during training, OVOD is typically formulated as a vision-language task, allowing detectors to leverage pretrained vision-language models (VLMs) for knowledge transfer from seen to unseen categories.
Existing methods mainly fall into two categories. The first aligns detection heads with pretrained VLMs, most notably CLIP~\cite{radford2021learning}, to exploit its strong zero-shot transfer ability by integrating CLIP features or aligning region features with text embeddings~\cite{zareian2021open,zhao2022exploiting,bangalath2022bridging,wu2023cora,wu2023aligning,wang2024sia,kim2024retrieval,wang2025ov}. While effective, these methods rely on CLIP's image-level pretraining, limiting robustness under distribution shifts.
The second line of work constructs object-aware vision-language spaces through large-scale heterogeneous supervision across classification~\cite{deng2009imagenet}, detection~\cite{lin2014microsoft,gupta2019lvis,shao2019objects365}, grounding~\cite{plummer2015flickr30k,yu2016modeling}, and image-text datasets~\cite{sharma2018conceptual}, enabling finer-grained cross-modal representations~\cite{zhong2022regionclip,li2022grounded,zhou2022detecting}.
In this work, we focus on the first paradigm, which has been widely adopted in recent OVOD approaches, and systematically investigate its robustness under distribution shifts and corrupted inputs.

\subsection{Domain Generalization}
Domain generalization (DG) aims to enable models to perform reliably across unseen or distribution-shifted domains without access to target-domain data during training~\cite{gulrajani2020search, wang2022generalizing}.
In computer vision, models often face diverse domain shifts caused by adverse weather, motion blur, sensor noise~\cite{hendrycks2019benchmarking, michaelis2019benchmarking,liu2021towards}, or style variations such as artistic renditions and synthetic-to-real transitions~\cite{mao2023coco, huang2021fsdr}.
Efforts to improve domain generalization can be broadly categorized into three directions.
Data-centric approaches enhance model robustness by synthesizing diverse domains or applying corruption- and style-based augmentations~\cite{hendrycks2019augmix, rebuffi2021fixing, cubuk2019autoaugment}.
Model-centric approaches focus on architectural or adaptation strategies, such as domain adversarial training, normalization modulation, and meta-learning~\cite{ganin2016domain, seo2020learning, chen2019progressive}.
Representation-centric approaches leverage self-supervised or contrastive pretraining to learn domain-invariant features, often preserving semantic consistency through vision-language alignment~\cite{radford2021learning, huang2024noise}.

Most studies focus on closed-vocabulary detection.
In contrast, open-vocabulary detection relies on fine-grained image-text alignment to generalize beyond seen categories.
Existing domain generalization and robustness approaches mainly enhance visual representations, yet they often overlook the cross-modal semantic consistency crucial for OVOD under domain shifts.

\subsection{Curriculum Learning}
Curriculum Learning (CL)~\cite{bengio2009curriculum} is a training paradigm that mimics the human learning process by organizing samples from easy to hard, thereby improving convergence and generalization. Extensive research has explored diverse formulations and scheduling strategies, including manually defined curricula and automatic mechanisms such as self-paced learning~\cite{kumar2010self}, teacher-student frameworks~\cite{graves2017automated}, and reinforcement-based sample selection~\cite{narvekar2020curriculum}. These methods generally consist of a difficulty estimator and a scheduler that jointly determine the learning order of training instances.
In computer vision, CL has been applied to domain adaptation, semi-supervised learning, and robustness enhancement under distribution shifts. In particular, curriculum strategies have proven effective for domain adaptation, \eg, C-SFDA~\cite{karim2023csfda} for source-free adaptation and curriculum domain adaptation for semantic segmentation~\cite{zhang2017curriculum}, both supporting the principle of exploiting reliable training signals first and gradually introducing harder or more ambiguous samples. Recent extensions further target multi-modal and cross-domain settings, including vision-language pretraining~\cite{zhou2022detecting, srinivasan2023curriculum} and robustness~\cite{cai2018curriculum}.

Most existing approaches overlook the progressive alignment between modalities and domains, focusing only on single-modality or data-level difficulty.
We introduce a curriculum-guided modality learning framework for domain-generalized open-vocabulary object detection, achieving a gradual refinement from coarse to fine-grained cross-modal alignment.

\begin{figure*}[!t]
  \centering
  \includegraphics[width=\linewidth]{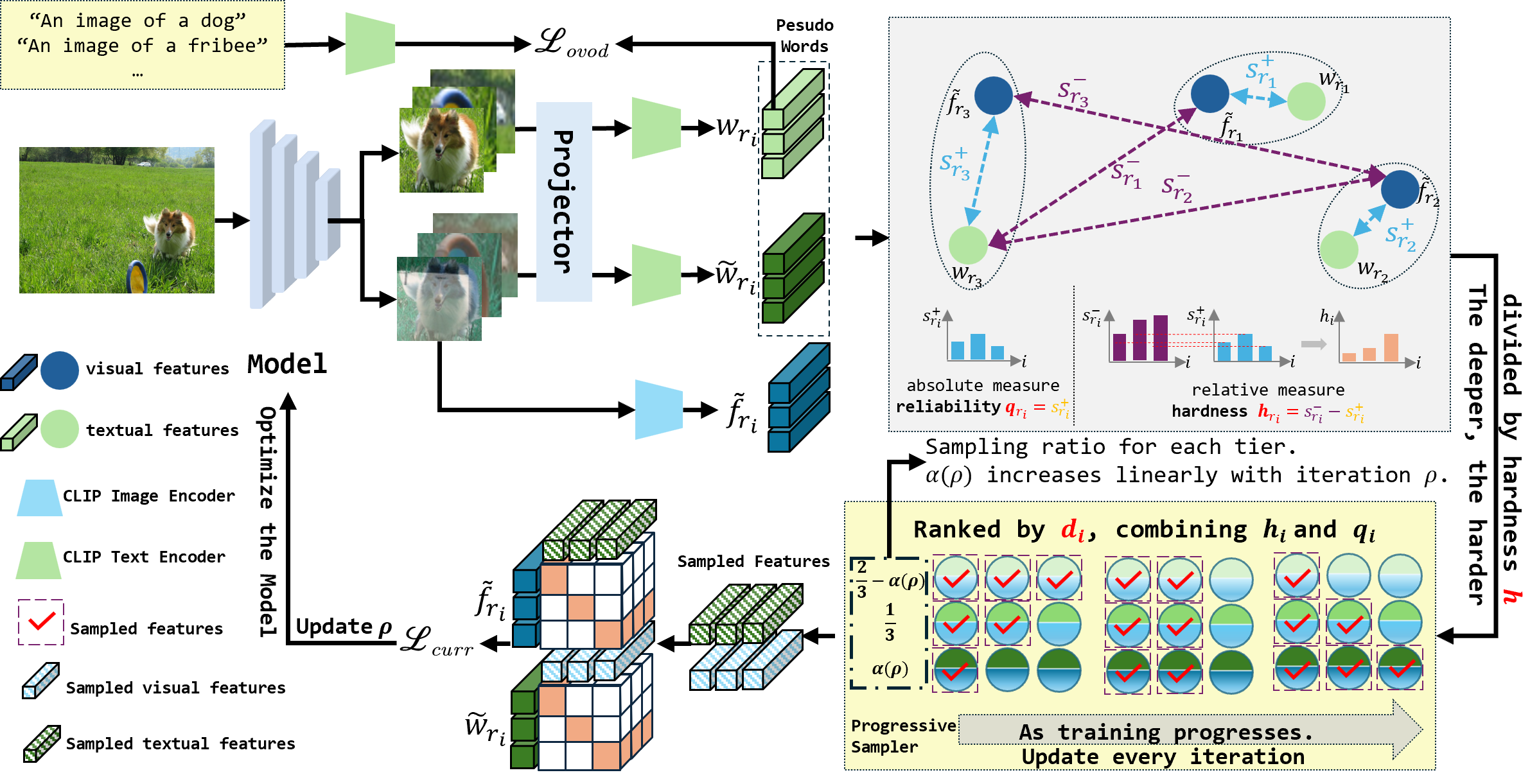}
  \caption{Overview of the Progressive Domain-invariant Cross-modal Alignment (PICA). The progressive sampler combines ambiguity proxy $h$ and signal strength proxy $q$ into a quality-adjusted difficulty score, then divides region pairs into easy, medium, and hard tiers. It uses a dynamic sampling ratio $\alpha(\rho)$, where $\rho$ represents the training iteration, to align selected region features with pseudo-word prototypes in a staged curriculum.}
  \label{fig:curr}
\end{figure*}

\section{Methodology}
\subsection{Problem Statement}
Open-Vocabulary Object Detection (OVOD) aims to detect objects from both base categories $\mathcal{C}_{\text{base}}$ and novel categories $\mathcal{C}_{\text{novel}}$, where $\mathcal{C}_{\text{base}} \cap \mathcal{C}_{\text{novel}} = \emptyset$. The model leverages a pretrained vision-language model (VLM)~\cite{radford2021learning}, with text and visual encoders denoted as $\mathscr{F}_{\text{text}}$ and $\mathscr{F}_{\text{vision}}$, respectively. For a class $c$, its semantic embedding is $t_c = \mathscr{F}_{\text{text}}(\mathbf{prompt}(c))$; for an image region $r$, its visual embedding is $f_r = \mathscr{F}_{\text{vision}}(r)$. Detection scores are computed as $s_{r,c} = \mathbf{sim}(f_r, t_c)$, with higher values indicating stronger cross-modal alignment.

For regions without associated class labels (\ie, from $\mathcal{C}_{\text{novel}}$), we employ a learnable pseudo-word prototype $w_r \in \mathbb{R}^d$ as a textual proxy that approximates the semantic content of region $r$:
\begin{equation}
    w_r = \mathscr{P}_{\text{img}\rightarrow\text{text}}(f_r) = \mathbf{W} f_r + \mathbf{b},
\end{equation}
where $\mathscr{P}_{\text{img}\rightarrow\text{text}}$ is a trainable projection head, and $\mathbf{W}, \mathbf{b}$ are learnable parameters, following common practice in vision-language alignment~\cite{minderer2022simple,radford2021learning,wu2023aligning,jia2021scaling}.
Crucially, $\mathscr{P}_{\text{img}\rightarrow\text{text}}$ is jointly optimized by $\mathcal{L}_{\text{ovod}}$, which grounds $w_r$ toward text embeddings $t_c$, ensuring that $w_r$ approximates a semantic position in the language embedding space rather than a self-referential projection artifact.

Standard OVOD typically assumes visual-domain stationarity between the source data $\mathcal{D}_{\text{src}}$ and target data $\mathcal{D}_{\text{tgt}}$, despite their category-space difference.
We extend this to Domain-Generalized Open-Vocabulary Object Detection~(DG-OVOD), where the model is trained on $\mathcal{D}_{\text{src}} = \{ (I^s, \mathcal{B}^s) \}$ with annotations only from $\mathcal{C}_{\text{base}}$, and evaluated on unseen $\mathcal{D}_{\text{tgt}} = \{ (I^t, \mathcal{B}^t) \}$ covering $\mathcal{C}_{\text{novel}}$, where $I^t$ exhibits joint distribution shift encompassing both semantic shift and visual divergence from $I^s$.

Under the DG-OVOD setting, the core challenge lies in the domain-induced variance of visual embeddings $f_r$. This variance can trigger alignment drift and cause substantial performance degradation on novel categories, which lack explicit supervision. A mechanism for improving domain-invariant cross-modal alignment is therefore needed.

To empirically examine this alignment-drift hypothesis, we evaluate a broad set of representative OVOD methods under our DG-OVOD evaluation benchmark, detailed in Table~\ref{table:coco-c} and Table~\ref{table:coco-o}.
The results show that rankings under clean conditions do not always predict robustness under shift. This evidence suggests that current alignment mechanisms, while effective in stationary environments, may lack sufficiently stable cross-domain correspondence to anchor novel-class signals against visual perturbations. This limitation motivates the progressive alignment approach developed in the following section.

\subsection{Sample-Adaptive Stratification under Domain Shift}
Domain shifts cause two distinct failure modes at the sample level, as illustrated in Fig.~\ref{fig:compare}. The first is signal collapse: the positive alignment score degrades, and the region feature weakly activates its matched pseudo-word prototype. The second is boundary confusion: the margin between the positive prototype and the hardest negative shrinks, and the region becomes close to a confounding semantic alternative.

These two failure modes are related but not interchangeable. Signal collapse reflects the absolute strength of the positive cross-modal match, whereas boundary confusion reflects the relative margin between the positive match and the strongest negative. A region may have a strong positive response but still be close to a hard negative, or it may have a weak positive response without being strongly confused with any particular negative.

An overview of PICA is shown in Fig.~\ref{fig:curr}. Accordingly, we use two complementary training-time proxies to estimate region-level alignment reliability: the \textbf{signal strength proxy} and the \textbf{ambiguity proxy}. These proxies are not intended to directly certify semantic correctness. Instead, they provide an operational ranking signal for selecting region pairs whose auxiliary cross-modal associations are less likely to be dominated by signal collapse or boundary confusion.
Specifically, for each region $r_i$ in a training image $I \in D_{src}$, we extract its visual feature $f_{r_i} = \mathscr{F}_{\text{vision}}(r_i)$. We also generate an augmented view $\tilde{I} = \mathcal{A}(I)$ and obtain the corresponding feature $\tilde{f}_{r_i} = \mathscr{F}_{\text{vision}}(r_i; \tilde{I})$. Each visual feature is mapped to the textual embedding space via the cross-modal projection $\mathscr{P}_{\text{img}\rightarrow\text{text}}$, yielding pseudo-word prototypes:
\begin{equation}
w_{r_i} = \mathscr{P}_{\text{img}\rightarrow\text{text}}(f_{r_i}), \quad
\tilde{w}_{r_i} = \mathscr{P}_{\text{img}\rightarrow\text{text}}(\tilde{f}_{r_i}).
\end{equation}

Let $s^+_{r_i}$ denote the positive similarity between the region feature and its matched pseudo-word prototype, and let $s^-_{r_i}$ represent the strongest confounding negative similarity from a candidate set $\mathcal{Z}^-$. The candidate set contains in-batch pseudo-word prototypes and cross-batch memory samples. Since the projection head is also optimized by $\mathcal{L}_{\text{ovod}}$, which aligns pseudo-word prototypes with class text embeddings, these negatives provide a practical estimate of region-level alignment ambiguity:
\begin{equation}
\begin{aligned}
s^+_{r_i} &= \mathbf{sim}(f_{r_i}, w_{r_i}), \\
s^-_{r_i} &= \max_{z \in \mathcal{Z}^-} \mathbf{sim}(f_{r_i}, z),
\end{aligned}
\end{equation}
where $\mathbf{sim}(\cdot,\cdot)$ denotes cosine similarity. We define the \textbf{signal strength proxy} $q_{r_i}$ as
\begin{equation}
q_{r_i} \triangleq s^+_{r_i}.
\end{equation}
A larger $q_{r_i}$ indicates a stronger activation of the matched pseudo-word prototype. Low-$q_{r_i}$ regions are therefore treated as less reliable candidates for the auxiliary alignment loss, because their positive cross-modal signal may be weak under noise, occlusion, or background dominance.

We define the \textbf{ambiguity proxy} $h_{r_i}$ as a margin-based proxy:
\begin{equation}
h_{r_i} \triangleq s^-_{r_i} - s^+_{r_i}.
\end{equation}
Larger values of $h_{r_i}$ indicate a smaller positive--negative margin $s^+_{r_i}-s^-_{r_i}$, meaning that the strongest negative is close to, or stronger than, the positive similarity. Such samples are useful for refining discriminative structure, but introducing too many of them in early training may destabilize the auxiliary alignment objective.

Guided by this perspective, PICA converts the two proxies into a single quality-adjusted difficulty score for curriculum sampling. After Z-score normalization, we define a low-signal indicator
\begin{equation}
\ell_i = \mathbb{I}\left[\hat{q}_i \le P_{p_q}(\hat{q})\right],
\end{equation}
where $\mathbb{I}[\cdot]$ denotes the indicator function, which equals $1$ when the enclosed condition is satisfied and $0$ otherwise, and $P_{p_q}(\hat{q})$ denotes the bottom-$p_q$ percentile of the normalized signal strengths in the current mini-batch. The final sampling difficulty is defined as
\begin{equation}
d_i = \hat{h}_i + \delta \ell_i.
\end{equation}
Here, $\hat{h}_i$ gives the main ambiguity-based ordering, while $\delta\ell_i$ adds a soft penalty to samples whose positive alignment signal falls into the bottom-$p_q$ percentile of the current mini-batch. This adjustment is useful because a small value of $h$ does not always mean that a region is reliable: both the positive and negative similarities can be low for weakly aligned or background-dominated regions. The indicator therefore serves as a reliability gate, pushing such low-signal regions to harder tiers with a tunable penalty $\delta$.
Regions with high ambiguity or weak positive signal are less likely to dominate the auxiliary loss at the beginning of training, but they can still be introduced as the curriculum expands toward harder samples. Let $\mathcal{R}$ denote the final set of $M$ selected regions. The complete procedure is summarized in Algorithm~\ref{alg:curriculum_tier_sampling}.

\begin{algorithm}[!t]
\caption{Progressive Curriculum Sampler for PICA}
\label{alg:curriculum_tier_sampling}
\begin{algorithmic}[1]
\REQUIRE Ambiguity proxy $h$; signal strength proxy $q$; Z-score normalization $z(\cdot)$; progress $\rho \in [0,1]$; total sample count $M$; base ratios $(r_E^0, r_M^0, r_H^0)$ s.t. $\sum r^0 = 1$; quality percentile threshold $p_q$ (default $0.05$); low-signal penalty $\delta$ (default $2.0$); scheduling function $\alpha(\rho)$ (default $\frac{2}{3}\rho$)
\ENSURE Sampled region set $\mathcal{R}$
\STATE Normalize scores: $\hat{h} \leftarrow z(h)$, $\hat{q} \leftarrow z(q)$
\STATE Compute low-signal indicators: $\ell_i \leftarrow \mathbb{I}[\hat{q}_i \le P_{p_q}(\hat{q})]$
\STATE Compute quality-adjusted difficulty scores: $d_i \leftarrow \hat{h}_i + \delta\ell_i$
\STATE Partition regions by ascending dynamic quantiles of $d$: $\mathcal{Q}_{Easy}$ contains the lowest-$d$ regions, $\mathcal{Q}_{Medium}$ contains intermediate regions, and $\mathcal{Q}_{Hard}$ contains the highest-$d$ regions
\STATE Compute target ratios with schedule $\alpha(\rho) \ge 0$: $\tilde{r}_{Easy} \leftarrow \max(0, r_E^0 - \alpha(\rho))$; $\tilde{r}_{Medium} \leftarrow r_M^0$; $\tilde{r}_{Hard} \leftarrow r_H^0 + \alpha(\rho)$
\STATE Normalize ratios: $(r_{Easy}, r_{Medium}, r_{Hard}) \leftarrow \text{Normalize}(\tilde{r}_{Easy}, \tilde{r}_{Medium}, \tilde{r}_{Hard})$
\STATE Determine per-tier sample counts: $n_t \leftarrow \lfloor M \cdot r_t \rfloor$, $\forall t \in \{Easy,Medium,Hard\}$
\FOR{each tier $t \in \{Easy,Medium,Hard\}$}
    \IF{$|\mathcal{Q}_t| > n_t$}
        \STATE Randomly sample $n_t$ items from $\mathcal{Q}_t$: $\mathcal{S}_t \leftarrow \text{RandomSample}(\mathcal{Q}_t, n_t)$
    \ELSE
        \STATE $\mathcal{S}_t \leftarrow \mathcal{Q}_t$ \COMMENT{Fallback if the tier is small}
    \ENDIF
\ENDFOR
\STATE Combine sampled regions: $\mathcal{R} \leftarrow \mathcal{S}_{Easy} \cup \mathcal{S}_{Medium} \cup \mathcal{S}_{Hard}$
\RETURN $\mathcal{R}$
\end{algorithmic}
\end{algorithm}

\begin{algorithm}[!t]
\caption{Full PICA Training Pipeline}
\label{alg:full_pica_training}
\begin{algorithmic}[1]
\REQUIRE Mini-batch $\{I_b,\mathcal{B}_b\}_{b=1}^{B}$ from $\mathcal{D}_{src}$; detector $\mathcal{G}$; frozen VLM encoders $\mathscr{F}_{\text{vision}},\mathscr{F}_{\text{text}}$; projection head $\mathscr{P}_{\text{img}\rightarrow\text{text}}$; memory queue $\mathcal{M}$; training step $t$ and total steps $T$
\ENSURE Updated detector and projection parameters
\STATE Compute progress $\rho \leftarrow t/T$ and generate augmented views $\tilde{I}_b=\mathcal{A}(I_b)$
\STATE Run $\mathcal{G}$ on each $I_b$ to obtain valid proposals/regions $\mathcal{V}_b$ and detection predictions
\STATE Extract paired region features $f_{r_i}$ and $\tilde{f}_{r_i}$ for all valid regions $r_i \in \mathcal{V}_b$
\STATE Project pseudo-word prototypes $w_{r_i}=\mathscr{P}_{\text{img}\rightarrow\text{text}}(f_{r_i})$ and $\tilde{w}_{r_i}=\mathscr{P}_{\text{img}\rightarrow\text{text}}(\tilde{f}_{r_i})$
\STATE Build candidate negatives $\mathcal{Z}^{-}$ from in-batch prototypes and memory queue $\mathcal{M}$
\STATE Dynamically compute $q_{r_i}$, $h_{r_i}$, and $d_i$ for every valid region in the mini-batch
\STATE Update sampled region set $\mathcal{R}$ by Algorithm~\ref{alg:curriculum_tier_sampling} with progress $\rho$
\STATE Compute $\mathcal{L}_{\text{det}}$ and $\mathcal{L}_{\text{ovod}}$ on all valid proposals/regions; compute $\mathcal{L}_{\text{curr}}$ only on $\mathcal{R}$
\STATE Optimize $\mathcal{L}_{\text{all}}=\mathcal{L}_{\text{det}}+\mathcal{L}_{\text{ovod}}+\lambda_{\text{curr}}\mathcal{L}_{\text{curr}}$
\STATE Enqueue current prototypes/features into $\mathcal{M}$ and dequeue stale entries
\end{algorithmic}
\end{algorithm}

\subsection{Curriculum-Guided Cross-Modal Alignment}
In DG-OVOD, curriculum learning is useful not because target-domain samples are known during training, but because the auxiliary cross-modal alignment signal has time-varying reliability. Early ambiguous or low-signal regions can produce unstable pseudo-word anchors, whereas low-difficulty regions provide a safer initialization for the projection space.

As detailed in Algorithm~\ref{alg:curriculum_tier_sampling} and Algorithm~\ref{alg:full_pica_training}, $q$, $h$, and $d$ are not precomputed dataset-level constants. They are recomputed for each mini-batch using the current projection head and memory queue. The sampled set $\mathcal{R}$ is therefore updated dynamically throughout training. Rather than applying the alignment regularizer uniformly to all regions, PICA first selects regions with low quality-adjusted difficulty. As training progresses, the hard-tier ratio increases and the auxiliary loss gradually includes samples with higher ambiguity or weaker signal. This schedule controls when unreliable cross-modal associations contribute to $\mathcal{L}_{\text{curr}}$, while preserving their later discriminative value after the projection becomes better calibrated.

The curriculum loss $\mathcal{L}_{\text{curr}}$ is computed only on the sampled set $\mathcal{R}$, while the detector loss $\mathcal{L}_{\text{det}}$ and the open-vocabulary loss $\mathcal{L}_{\text{ovod}}$ remain computed on all valid proposals/regions in the mini-batch. This separation is important: PICA changes the scope of the auxiliary alignment regularizer, not the detection supervision itself. Rare or hard regions that are not selected into $\mathcal{R}$ at an early iteration can still contribute to $\mathcal{L}_{\text{det}}$ and $\mathcal{L}_{\text{ovod}}$, and may enter $\mathcal{R}$ later as the schedule increases the hard-tier ratio.
The auxiliary loss is formulated as a bidirectional InfoNCE objective over $\mathcal{R}$:
\begin{equation}
\begin{aligned}
\mathcal{L}_{\text{curr}}
= \frac{1}{2|\mathcal{R}|}
\sum_{r_i \in \mathcal{R}}
\Big[&
\mathcal{N}\big(\tilde{f}_{r_i}, w_{r_i}, \{w_{r_j}\}_{r_j \in \mathcal{R}}\big) \\
&+ \mathcal{N}\big(\tilde{w}_{r_i}, \tilde{f}_{r_i}, \{\tilde{f}_{r_j}\}_{r_j \in \mathcal{R}}\big)
\Big],
\end{aligned}
\end{equation}
where $\mathcal{N}(a,p,\mathcal{P})$ denotes an InfoNCE term with anchor $a$, positive $p$, and candidate set $\mathcal{P}$:
\begin{equation}
\mathcal{N}(a,p,\mathcal{P}) = -\log \frac{\exp(\mathbf{sim}(a,p)/\tau)}{\sum_{u\in\mathcal{P}}\exp(\mathbf{sim}(a,u)/\tau)}.
\end{equation}
We adopt a learnable temperature parameter $\tau$ initialized to 0.07. This objective uses paired clean and augmented region features as a consistency regularizer. Since the pseudo-word prototypes are also constrained by $\mathcal{L}_{\text{ovod}}$, the loss encourages the projected prototypes to be less sensitive to local perturbations and latent-space drift.

\subsection{Overall Objective}
Overall, the complete training objective integrates detection loss $\mathcal{L}_{\text{det}}$~\cite{ren2015faster}, open-vocabulary loss $\mathcal{L}_{\text{ovod}}$~\cite{zareian2021open,wu2023aligning}, and our curriculum losses:
\begin{equation}
\mathcal{L}_{\text{all}} = \mathcal{L}_{\text{det}} + \mathcal{L}_{\text{ovod}} + \lambda_{\text{curr}}\mathcal{L}_{\text{curr}}.
\end{equation}

\begin{table*}[!t]
  \scriptsize
  \centering
  \caption{Evaluation of Recent OVOD Methods Across Diverse Corruption Types on OV-COCO-C. Each Result Reports Novel-class $\text{mAP}^{\text{dom}_k}_{50}$ ($\mathcal{L}_k=5$). $\text{mAP}^{\text{avg}}_{50}$ Denotes the Mean Across All 15 Corruption Types. ``Standard'' Corresponds to the Standard Evaluation Protocol on OV-COCO Without Corruption. }
  \label{table:coco-c}
  \setlength{\tabcolsep}{1.2pt}
  \renewcommand{\arraystretch}{1.3}
  \begin{tabularx}{\linewidth}{l | CCCCCCCCCC | C}
  \toprule
    \textbf{Exp.} &
    \scalebox{0.95}{R.CLIP~\cite{zhong2022regionclip}} &
    \scalebox{0.95}{VL-PLM~\cite{zhao2022exploiting}} &
    \scalebox{0.95}{Obj~\cite{bangalath2022bridging}} &
    \scalebox{0.95}{CORA~\cite{wu2023cora}} &
    \scalebox{0.95}{Baron~\cite{wu2023aligning}} &
    \scalebox{0.95}{SIA~\cite{wang2024sia}} &
    \scalebox{0.95}{RALF~\cite{kim2024retrieval}} &
    \scalebox{0.95}{DQUO~\cite{wang2025ov}} &
    \scalebox{0.95}{Baron$^{*}$~\cite{wu2023aligning}} & \scalebox{0.95}{Baron$^{\dag}$~\cite{wu2023aligning}} & \textbf{PICA} \\ \midrule
    Standard & 31.4 & 34.4 & 36.6 & 35.1 & 34.0 & 35.3 & 41.3 & 39.2 & 33.4 & 34.4 & {37.5} \\ \midrule
    Gauss    & 12.5 & 14.1 & 16.1 & 8.6  & 17.8 & 9.1  & 16.0 & 9.7  & 15.8 & 17.7 & 17.8 \\
    Shot     & 12.7 & 14.6 & 16.6 & 8.7  & 17.3 & 9.3  & 16.6 & 9.4  & 16.1 & 17.5 & 17.8 \\
    Impulse  & 10.7 & 10.3 & 11.5 & 5.4  & 13.6 & 5.7  & 11.5 & 6.6  & 12.3 & 14.4 & 13.8 \\
    Defocus  & 13.2 & 16.8 & 19.6 & 6.8  & 18.3 & 7.6  & 19.6 & 8.0  & 16.5 & 18.9 & 19.6 \\
    Glass    & 5.0  & 10.8 & 12.2 & 4.6  & 12.2 & 4.7  & 12.2 & 5.4  & 10.9 & 12.1 & 12.5 \\
    Motion   & 12.7 & 16.4 & 19.0 & 9.4  & 17.0 & 9.7  & 19.0 & 10.6 & 15.1 & 16.8 & 17.4 \\
    Zoom     & 7.8  & 10.8 & 11.4 & 5.4  & 10.7 & 5.6  & 11.4 & 6.0  & 9.2  & 10.7 & 11.0 \\
    Snow     & 12.9 & 15.7 & 17.1 & 11.4 & 17.6 & 11.9 & 17.1 & 13.0 & 17.4 & 18.9 & 19.8 \\
    Frost    & 17.2 & 19.7 & 22.3 & 13.7 & 21.4 & 16.4 & 22.4 & 18.3 & 20.9 & 22.3 & 22.8 \\
    Fog      & 25.6 & 26.2 & 30.8 & 27.0 & 29.2 & 27.7 & 31.2 & 30.5 & 28.3 & 30.1 & 31.3 \\
    Bright   & 26.0 & 29.2 & 34.8 & 28.8 & 30.8 & 29.2 & 34.8 & 32.3 & 30.2 & 31.3 & 33.6 \\
    Contrast & 23.3 & 18.6 & 22.1 & 19.1 & 23.0 & 20.2 & 22.1 & 21.8 & 21.8 & 24.3 & 24.7 \\
    Elastic  & 18.0 & 22.8 & 26.7 & 19.2 & 23.6 & 19.6 & 26.7 & 22.1 & 22.3 & 23.7 & 25.6 \\
    Pixelate & 12.3 & 14.4 & 16.3 & 10.8 & 17.3 & 11.2 & 16.3 & 12.4 & 17.4 & 17.8 & 19.5 \\
    JPEG     & 11.6 & 12.1 & 16.3 & 12.6 & 19.9 & 13.4 & 16.3 & 14.6 & 19.4 & 20.6 & 22.6 \\ \midrule
    \textbf{mAP$^{\text{avg}}_{50}$} & 14.8 & 16.8 & 19.5 & 12.8 & 19.3 & 13.4 & 19.5 & 14.7 & 18.3 & 19.8 & \textbf{20.7} \\ \bottomrule
  \end{tabularx}
  \vspace{2pt}
  \parbox{\linewidth}{\footnotesize \emph{Protocol note.} PICA, Baron, Baron$^{*}$, and Baron$^{\dag}$ use Faster R-CNN with ResNet-50-FPN and CLIP ViT-B/32 under the same OV-COCO split, 90k training schedule, and evaluation protocol; Baron$^{*}$ and Baron$^{\dag}$ add Mixup with DDC and DeepCORAL, respectively. Other baselines are evaluated using their official OV-COCO configurations under the same corruption protocol and are included only when their framework and available predictions/models are compatible with this evaluation.}
\end{table*}

\section{Experiment}
In this section, we present the experimental setup, datasets, implementation details, and evaluation of our proposed Progressive Domain-invariant Cross-modal Alignment (PICA) framework. We compare PICA with representative Open-Vocabulary Object Detection (OVOD) methods and conduct ablation studies to analyze the effect of each component.

\subsection{Experimental Setup}
\subsubsection{Datasets}
We follow the standard OV-COCO protocol~\cite{zareian2021open}, partitioning COCO-2017 into 48 base and 17 novel categories. Models are trained exclusively on source-domain images with base-class annotations and are evaluated on the 17 novel categories. 
For robustness evaluation, we construct two evaluation settings following~\cite{chhipa2024open}. \textbf{OV-COCO-C} applies 15 corruption types~\cite{hendrycks2019benchmarking} to the OV-COCO evaluation images, with five severity levels for each corruption type. For natural OOD evaluation, we use COCO-O~\cite{mao2023coco}, an existing benchmark for object detection under natural distribution shifts, and derive \textbf{OV-COCO-O} by aligning its six OOD domains with the standard OV-COCO 48/17 base-novel split. Specifically, we keep images and annotations whose categories belong to the 65 OV-COCO categories and discard categories outside this vocabulary. During evaluation, we report novel-class performance on the 17 OV-COCO novel categories under each OOD domain. No target-domain images or annotations from OV-COCO-C or OV-COCO-O are used during training. Together, these settings provide a unified assessment of open-vocabulary generalization across both synthetic and real-world domain shifts. The detailed dataset construction is shown in the Appendix.

\subsubsection{Evaluation Metrics}
Following standard practice in open-vocabulary object detection~\cite{zareian2021open,zhao2022exploiting,bangalath2022bridging,wu2023cora,wu2023aligning,wang2024sia,kim2024retrieval,wang2025ov}, we adopt $\text{mAP}_{50}$ as the primary evaluation metric.
Our experiments involve $K$ target domains $\mathcal{D}_{\text{tgt}} = \{\mathcal{D}_1, \dots, \mathcal{D}_K\}$, each containing multiple severity levels $\mathcal{L}_k$ for visual perturbations. To aggregate results, we first compute the mean $\text{mAP}_{50}$ within each domain:
\begin{equation}
\text{mAP}^{\text{dom}_k}_{50} = \frac{1}{\mathcal{L}_k} \sum_{\ell=1}^{\mathcal{L}_k} \text{mAP}_{50}(\mathcal{D}_k^{(\ell)}),
\label{eq:metrics_domain}
\end{equation}
where $\mathcal{D}_k^{(\ell)}$ denotes the subset of domain $\mathcal{D}_k$ at severity level $\ell$.
The overall performance is then obtained by averaging across all domains:
\begin{equation}
\text{mAP}^{\text{avg}}_{50} = \frac{1}{K} \sum_{k=1}^{K} \text{mAP}^{\text{dom}_k}_{50}.
\label{eq:metrics_all}
\end{equation}
This hierarchical averaging ensures that both severity-level variations and cross-domain differences are fairly reflected in the final evaluation score.

\subsubsection{Implementation Details}
We train a two-stage OVOD model: Faster-RCNN~\cite{ren2015faster} with a ResNet-50-FPN backbone and CLIP ViT-B/32 guidance on 4$\times$RTX 5090 GPUs for 90k iterations.
Training uses SGD with momentum 0.9, weight decay $2.5\times10^{-5}$, and an initial learning rate of 0.04 with linear warmup and multi-step decay.
Mixed precision is enabled via AmpOptimWrapper, and gradients are accumulated every two iterations, resulting in an effective batch size of 64.
MixUp augmentation is applied to the OVOD branch to enhance cross-modal diversity and improve generalization.
Both image and text features are divided into three tiers, with $M=4096$ samples per modality. $\lambda_{curr}=1.0$.
The curriculum sampler, memory queue, and auxiliary alignment loss are used only during training. At inference time, PICA uses the same detector architecture as the matched Baron baseline and introduces no additional sampling or memory module.
Following Baron~\cite{wu2023aligning}, the total number of iterations $T$ is 90k.
For fair comparison, Baron~\cite{wu2023aligning} is used as the primary baseline under identical detector, backbone, VLM guidance, training split, and schedule.

\subsection{Comparison with Other Methods}
\label{sec:comparison}
We evaluate our approach on OV-COCO-C and OV-COCO-O against recent OVOD methods.
For a matched comparison, we also integrate classic data augmentation~\cite{zhang2017mixup} and classic domain adaptation techniques, DDC~\cite{tzeng2014deep} and DeepCORAL~\cite{sun2016deep}, into Baron~\cite{wu2023aligning} as additional baselines for OV-COCO-C.
Since recent OVOD methods are released with different model designs and training configurations, the matched Baron comparison serves as the main controlled comparison. The remaining methods place the results in context under the same DG-OVOD benchmark.

\subsubsection{Performance under Corruptions}
Table~\ref{table:coco-c} reports the results on OV-COCO-C. PICA reaches \textbf{20.7} in $\text{mAP}^{\text{avg}}_{50}$, giving the best average result under this benchmark.
Across corruption types, including low-level noise (Gaussian, Shot), blur (Defocus, Glass, Motion, Zoom), weather shifts (Snow, Frost, Fog, Bright), and photometric/geometric distortions (Contrast, Elastic, Pixelation, JPEG), PICA improves over the matched Baron baseline on most corruptions and improves the average over Baron$^{\dag}$. For instance, under Snow, PICA improves novel-class performance from 18.9 to 19.8, and under JPEG from 20.6 to 22.6. The gains are strongest at the aggregate level; some individual corruptions, such as Impulse and Bright, remain challenging.

\subsubsection{Performance under Out-of-Distributions Shift}
We further evaluate PICA on OV-COCO-O to assess domain generalization under natural OOD domains. As shown in Table~\ref{table:coco-o}, PICA achieves an average $\text{mAP}^{\text{avg}}_{50}$ of 19.7.
PICA improves over the matched Baron baseline on Cartoon, Handmade, Painting, and Weather, with notable gains under Painting and Weather. Performance degrades on Tattoo and remains nearly flat on Sketch, where extreme stylization and sparse structure can make the cross-modal alignment proxies less reliable. These results position PICA as a training-time stabilizer for projection-based OVOD, especially under corruptions and moderate appearance shifts.

\begin{table}[!t]
\centering
\scriptsize
\caption{Comparison of Recent OVOD Methods Under Various OOD Domains on OV-COCO-O. Results Report Novel-class $\text{mAP}^{\text{dom}_k}_{50}$ with $\mathcal{L}_k=1$.}
\label{table:coco-o}
\setlength{\tabcolsep}{3pt}
\renewcommand{\arraystretch}{1.3}
\begin{tabularx}{\linewidth}{l | CCCC | C}
\toprule
\textbf{Domain} & RegionCLIP~\cite{zhong2022regionclip} & Obj~\cite{bangalath2022bridging} & Baron~\cite{wu2023aligning} & RALF~\cite{kim2024retrieval} & {PICA} \\ \midrule
Cartoon     & 14.6 & 5.7  & 12.1 & 9.8  & 13.3 \\
Handmade    & 13.4 & 12.5 & 17.6 & 13.9 & 19.9 \\
Painting    & 15.2 & 17.2 & 23.4 & 19.5 & {29.5} \\
Sketch      & 13.2 & 7.4  & 8.9  & 8.5  & 8.8  \\
Tattoo      & 16.9 & 9.7  & 10.0 & 14.4 & 7.2  \\
Weather     & 25.8 & 22.7 & 37.1 & 28.4 & {39.3} \\ \midrule
\textbf{mAP$^{\text{avg}}_{50}$} & 16.5 & 12.5 & 18.5 & 15.8 & {19.7} \\ \bottomrule
\end{tabularx}
\vspace{2pt}
\parbox{\linewidth}{\footnotesize \emph{Protocol note.} This comparison reports only methods that can be evaluated without pre-extracted region embeddings under the OV-COCO-O split. Baron and PICA use the same Faster R-CNN/ResNet-50-FPN/CLIP ViT-B/32 training protocol; RegionCLIP, Obj, and RALF are reported with their compatible official settings for contextual comparison.}
\end{table}

\subsubsection{Generalization to Stronger Backbones and Real-World Conditions}
To examine whether the benefit of PICA is tied to a specific backbone or to synthetic corruptions, we conduct two additional studies, summarized in Table~\ref{tab:scaling_realworld}.
First, we scale the visual--language backbone from CLIP ViT-B/32 to the stronger CLIP ViT-L/14. PICA reaches \textbf{22.1} $\text{mAP}^{\text{avg}}_{50}$ on OV-COCO-C, improving over the baseline (21.1), which suggests that the curriculum-guided alignment remains useful as representational capacity grows.
Second, to move beyond synthetic corruptions and artistic OOD styles, we evaluate on Rainy-Cityscapes restricted to the COCO-overlapping categories, a realistic driving/adverse-weather setting. PICA also improves over the baseline (\textbf{42.4} vs.\ 40.2 $\text{mAP}^{\text{avg}}_{50}$), suggesting that the training-time alignment regularizer can also benefit a real-world weather shift.
We further note that PICA is stable across three random seeds on OV-COCO-C, obtaining $20.63 \pm 0.19$ novel-class $\text{mAP}^{\text{avg}}_{50}$.

\begin{table}[!t]
\centering
\footnotesize
\caption{Generalization to a Stronger Backbone (CLIP ViT-L/14 on OV-COCO-C) and to a Real-World Adverse-Weather Benchmark (Rainy-Cityscapes, COCO-overlapping Categories). All Results Report Novel-class $\text{mAP}^{\text{avg}}_{50}$.}
\label{tab:scaling_realworld}
\setlength{\tabcolsep}{8pt}
\renewcommand{\arraystretch}{1.2}
\begin{tabular}{ll|cc}
\toprule
\textbf{Setting} & \textbf{Benchmark} & \textbf{Baseline} & \textbf{PICA} \\
\midrule
CLIP ViT-L/14 & OV-COCO-C & 21.1 & \textbf{22.1} \\
ResNet-50-FPN & Rainy-Cityscapes & 40.2 & \textbf{42.4} \\
\bottomrule
\end{tabular}
\end{table}

\begin{table}[!t]
\centering
\scriptsize
\caption{Ablation Study of PICA Components and Curriculum Sampling Strategies on Both OV-COCO-C and OV-COCO-O. SAS Denotes Sample-Adaptive Stratification, and $h$ or $q$ Indicates Which Proxy Guides Curriculum-based Cross-modal Alignment. ``Std.'' Is the Standard Protocol on OV-COCO; ``C-Avg.'' and ``O-Avg.'' Are the $\text{mAP}^{\text{avg}}_{50}$ on OV-COCO-C and OV-COCO-O, Respectively. All Results Are Reported on Novel Categories. ``--'' Means the Configuration Is Not Reported in That Setting.}
\label{tab:unified_ablation}
\setlength{\tabcolsep}{4pt}
\renewcommand{\arraystretch}{1.2}
\begin{tabular}{cccc|ccc}
\toprule
SAS & Mixup & $h$ & $q$ & Std. & C-Avg. & O-Avg. \\
\midrule
\ding{55} & \ding{55} & - & - & 34.0 & 19.3 & 18.5 \\
\ding{55} & \ding{51} & - & - & 34.2 & 19.6 & 19.0 \\
\ding{51} & \ding{55} & \ding{51} & \ding{51} & 36.5 & 20.2 & 19.0 \\
\ding{51} & \ding{51} & \ding{51} & \ding{55} & 34.0 & 19.8 & 18.7 \\
\ding{51} & \ding{51} & \ding{55} & \ding{51} & 34.2 & 19.6 & 17.4 \\
\ding{51} & \ding{51} & \ding{51} & \ding{51} & \textbf{37.5} & \textbf{20.7} & \textbf{19.7} \\
\bottomrule
\end{tabular}
\end{table}

\begin{table}[!t]
\caption{Ablation of Pseudo-word Generation Strategies in PICA Loss on OV-COCO-C. All Results Are Reported on Novel Categories Using the $\text{mAP}^{\text{avg}}_{50}$.}
\label{tab:pseudo_word_ablation}
\centering
\footnotesize
\setlength{\tabcolsep}{6pt}
\renewcommand{\arraystretch}{1.2}
\begin{tabular}{c|c|c}
\toprule
Pseudo-word Features & \textbf{Standard} & \textbf{mAP$^{\text{avg}}_{50}$}  \\ \midrule
$w_r = \mathscr{P}(f_r)$ & 36.1 & 20.0 \\
$w_r = \mathscr{P}(\tilde{f}_r)$ & 36.0 & 20.1 \\
$w_r, \tilde{w}_r$ & \textbf{37.5}  & \textbf{20.7} \\ \bottomrule
\end{tabular}
\end{table}

\begin{table}[!t]
\centering
\footnotesize
\caption{Controlled Comparison of PICA's Dual-proxy Sampler with Single-scalar Uncertainty Samplers on OV-COCO-C. All Results Are Reported on Novel Categories Using the $\text{mAP}^{\text{avg}}_{50}$ Under the Same Training Protocol.}
\label{tab:single_proxy}
\setlength{\tabcolsep}{6pt}
\renewcommand{\arraystretch}{1.2}
\begin{tabular}{l|c}
\toprule
\textbf{Sampler} & \textbf{mAP$^{\text{avg}}_{50}$} \\ \midrule
Entropy-only                  & 19.82 \\
Margin-based                  & 20.12 \\
\textbf{PICA (dual-proxy $q$\,\&\,$h$)} & \textbf{20.7} \\ \bottomrule
\end{tabular}
\end{table}

\subsection{Ablation Study}
To better understand the contribution of each component in PICA, we conduct ablation studies on both the synthetic OV-COCO-C and the natural OOD OV-COCO-O evaluation settings. Specifically, we analyze the effects of (1) Sample-Adaptive Stratification and (2) Alignment Invariance Learning, including pseudo-word generation and Mixup augmentation.
As summarized in Table~\ref{tab:unified_ablation}, the full PICA model, integrating sample-adaptive stratification, cross-modal consistency learning, and Mixup, achieves the highest $\text{mAP}^{\text{avg}}_{50}$ on \emph{both} evaluation settings. This result indicates that these components contribute complementarily to DG-OVOD.

\noindent \textbf{Effect of Sample-Adaptive Stratification.}
Table~\ref{tab:unified_ablation} shows the benefit of our curriculum mechanism. Replacing sample-adaptive stratification with a fixed feature queue reduces novel-category performance, which is consistent with the role of progressive sample-adaptive ranking in stabilizing cross-modal alignment under domain shifts. The Mixup-only variant improves the baseline from 19.3 to 19.6 on OV-COCO-C and from 18.5 to 19.0 on OV-COCO-O. By comparison, sample-adaptive stratification without Mixup reaches 20.2 on OV-COCO-C. This separates the contribution of the curriculum-guided alignment mechanism from generic augmentation. Using either ambiguity $h$ or signal strength $q$ in isolation also leads to inferior results, supporting the dual-proxy design.

\noindent \textbf{Consistency across Synthetic and Natural Shifts.}
Table~\ref{tab:unified_ablation} shows a similar component ranking across the two evaluation settings. On OV-COCO-O, the full design reaches 19.7 $\text{mAP}^{\text{avg}}_{50}$, above the no-SAS baseline (18.5), the Mixup-only variant (19.0), and the single-proxy variants ($h$-only 18.7, $q$-only 17.4). This mirrors the trend on OV-COCO-C and supports the contribution of the dual-proxy curriculum under both synthetic corruptions and natural distribution shifts.

\noindent \textbf{Effect of Cross-modal Alignment.}
We further investigate the role of pseudo-word prototypes in encouraging cross-modal consistency. Table~\ref{tab:pseudo_word_ablation} reports the effect of generating pseudo-words using only clean features, only augmented features, or both. Keeping all other components unchanged, using both clean and augmented features yields the best cross-domain performance, while using only one type reduces performance.

\noindent \textbf{Comparison with Single-Proxy Samplers.}
We further contrast our dual-proxy sampler with two representative single-scalar uncertainty samplers under the same OV-COCO-C benchmark, as shown in Table~\ref{tab:single_proxy}. The entropy-only and margin-based samplers obtain 19.82 and 20.12 novel-class $\text{mAP}^{\text{avg}}_{50}$, respectively, below PICA's 20.7. This result indicates that explicitly separating signal strength $q$ from boundary ambiguity $h$ is more effective than collapsing alignment difficulty into a single uncertainty score.
In our design, $q$ and $h$ are training-time operational proxies used to rank auxiliary region--prototype pairs during optimization. Their utility is evaluated through controlled training variants, including the $h$-only, $q$-only, dual-proxy, entropy-only, and margin-based samplers.

\subsection{Analysis}
\label{sec:analysis}
To further examine whether PICA is associated with more stable cross-modal behavior, we conduct quantitative analyses.
\begin{figure*}[!t]
  \centering
  \subfloat[]{\includegraphics[width=0.49\linewidth]{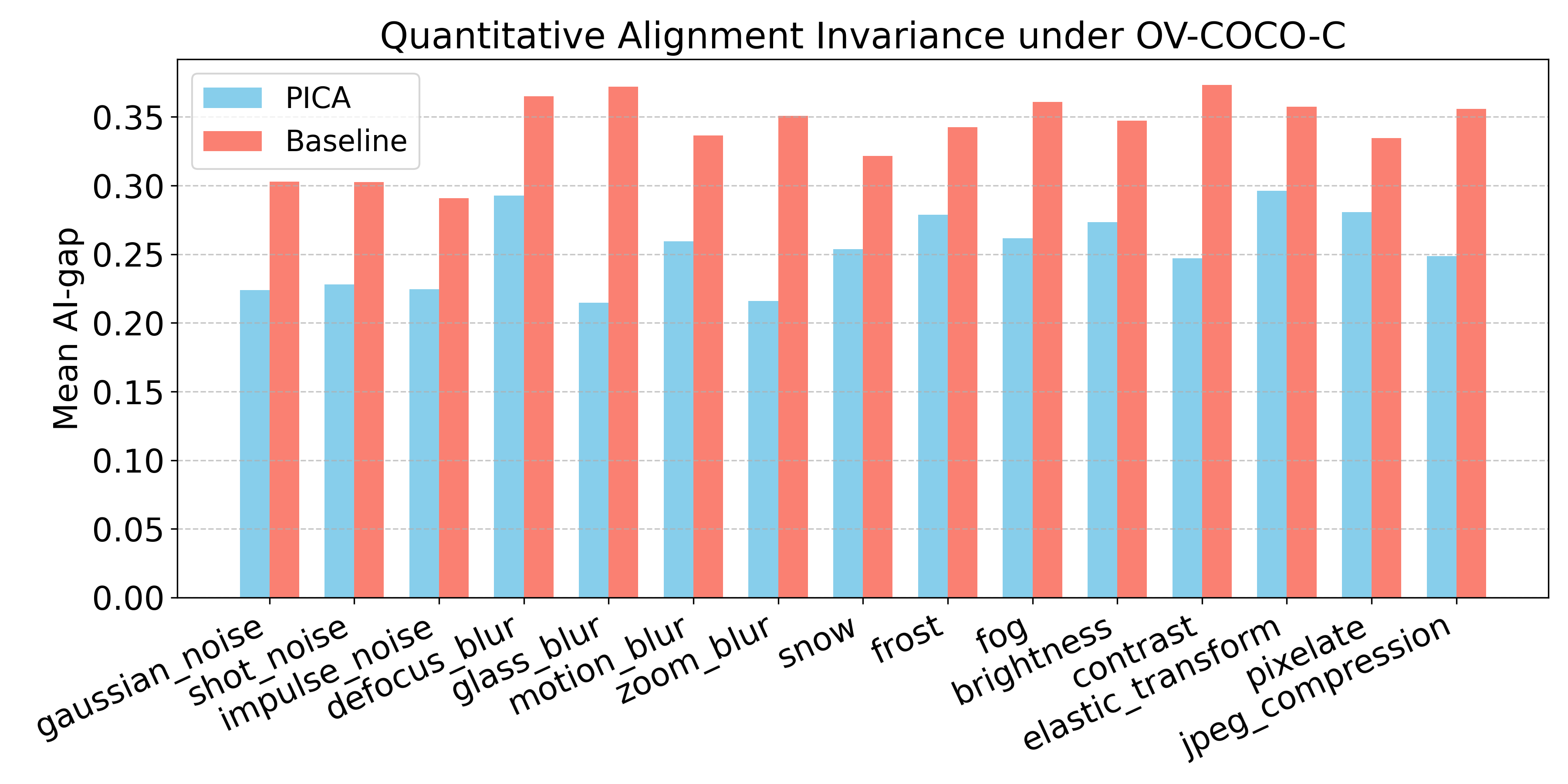}%
  \label{fig:mi_gap}}
  \hfil
  \subfloat[]{\includegraphics[width=0.49\linewidth]{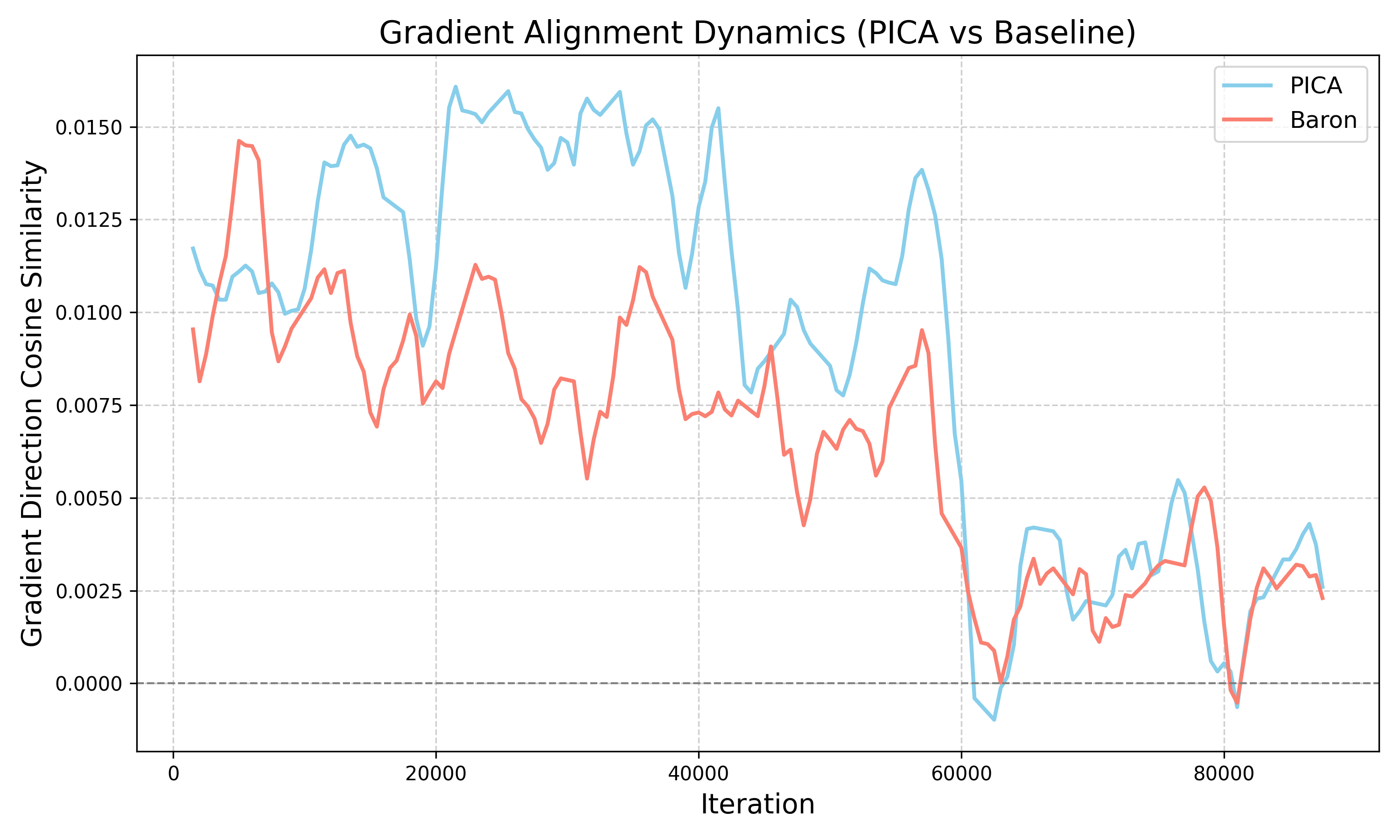}%
  \label{fig:grad_alignment}}
  \caption{Comparative analysis of cross-modal alignment and training stability. (a) Mean AI-gap for PICA and Baseline across OV-COCO-C. Lower AI-gap indicates more stable cross-domain alignment. (b) Gradient direction cosine similarity during training. PICA maintains smoother and higher alignment consistency than Baseline, which is consistent with reduced gradient conflicts and more stable cross-modal learning.}
  \label{fig:short}
\end{figure*}

\begin{figure*}[!t]
  \centering
  \subfloat[]{\includegraphics[width=0.49\linewidth]{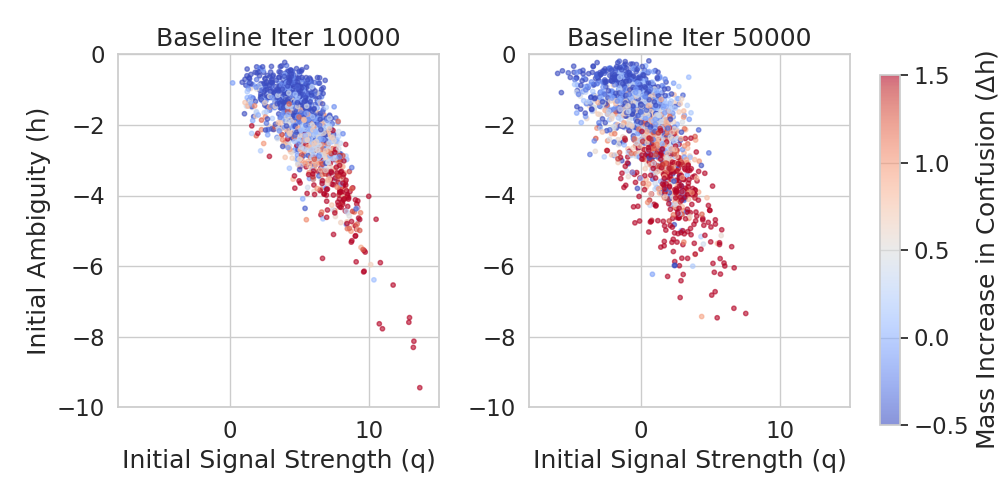}%
  \label{fig:baseline_checkpoints_stability}}
  \hfil
  \subfloat[]{\includegraphics[width=0.49\linewidth]{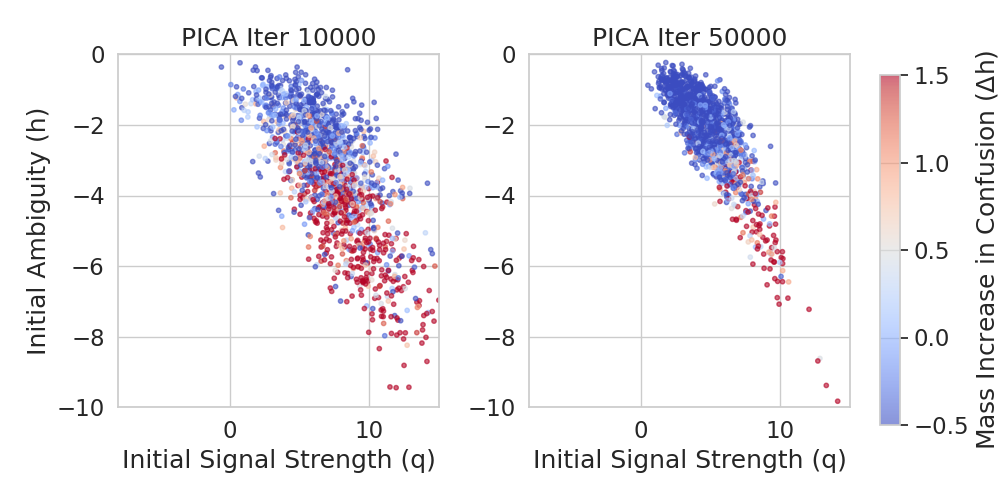}%
  \label{fig:pica_checkpoints_stability}}
  \caption{Region-level cross-modal alignment stability between standard and Gaussian noise. Each point represents a region, plotted by its clean-image signal strength $q$ (x-axis) and ambiguity $h$ (y-axis), colored by confusion increase $\Delta h = h_{\text{corrupted}} - h_{\text{clean}}$. (a) Baseline exhibits persistent high-$\Delta h$ dispersion across mid-to-high signal levels. (b) PICA progressively stabilizes mid-to-high signal regions ($\Delta h \to 0$), suggesting that curriculum training selectively consolidates alignment robustness for reliable regions.}
  \label{fig:stability}
\end{figure*}

\noindent \textbf{Quantitative Evaluation of Cross-modal Alignment Stability.}
We quantify cross-modal alignment stability under diverse domains using the Alignment Invariance Gap (AI-gap):
\begin{equation}
\text{AI-gap}_{k} = 1 - \text{sim}\big[\text{sim}(f_r^{clean}, t_c),\text{sim}(f_r^{k}, t_c)\big],
\end{equation}
where $k \in \mathcal{D}_{tgt}$ denotes the target domain, $\text{sim}(\cdot)$ is the cosine similarity, $f_r^{clean}$ and $f_r^{k}$ are region features from clean and target domain, $\text{sim}(f_r^{clean}, t_c)$ and $\text{sim}(f_r^{k}, t_c)$ are logit vectors $\in \mathbb{R}^C$, and $t_c$ is the textual embedding of category $c$.
Smaller values of $\text{AI-gap}_k$ indicate more stable visual-textual correspondence across domains, as illustrated in Fig.~\ref{fig:mi_gap}.

\noindent \textbf{Gradient Alignment Dynamics.}
To analyze the optimization behavior associated with PICA's improved robustness over Baseline, we examine the evolution of the gradient direction cosine similarity (GCS) between consecutive training steps:
\begin{equation}
\text{GCS}(\mathbf{g}_t, \mathbf{g}_{t-1}) =
\frac{\mathbf{g}_t \cdot \mathbf{g}_{t-1}}{\|\mathbf{g}_t\|\|\mathbf{g}_{t-1}\|},
\end{equation}
where $\mathbf{g}_t$ and $\mathbf{g}_{t-1}$ denote the full gradient at iterations $t$ and $t-1$. Higher GCS values indicate more consistent update directions, which are associated with stable optimization and better generalization~\cite{keskar2016large,wang2025gradient,liu2021conflict}.
As shown in Fig.~\ref{fig:grad_alignment}, PICA maintains higher GCS with stronger directional consistency and smaller gradient fluctuations during the first 60,000 iterations. While Baseline shows frequent drops that indicate gradient conflicts, PICA's curriculum gradually introduces samples from easy to hard. This pattern is consistent with fewer abrupt gradient conflicts and more stable optimization.

\noindent \textbf{Region-level Cross-modal Alignment Stability Analysis.}
To examine how curriculum training is associated with region-level alignment under distribution shift, we compute the confusion increase $\Delta h = h_{\text{corrupted}} - h_{\text{clean}}$ between standard and Gaussian noise. As illustrated in Fig.~\ref{fig:stability}, the baseline exhibits persistent and widespread high-$\Delta h$ across the entire signal spectrum ($q$) at both checkpoints, suggesting that uniform training is less selective when handling reliable and unreliable regions. In contrast, PICA shows a more structured stability pattern: by Iter 50000, mid-to-high signal regions tend to stabilize ($\Delta h \to 0$), reducing the impact of domain noise. This supports the interpretation that curriculum training can consolidate alignment robustness for informative regions without claiming to resolve all hard or rare cases.

\begin{figure}[!t]
  \centering
  \includegraphics[width=0.75\linewidth]{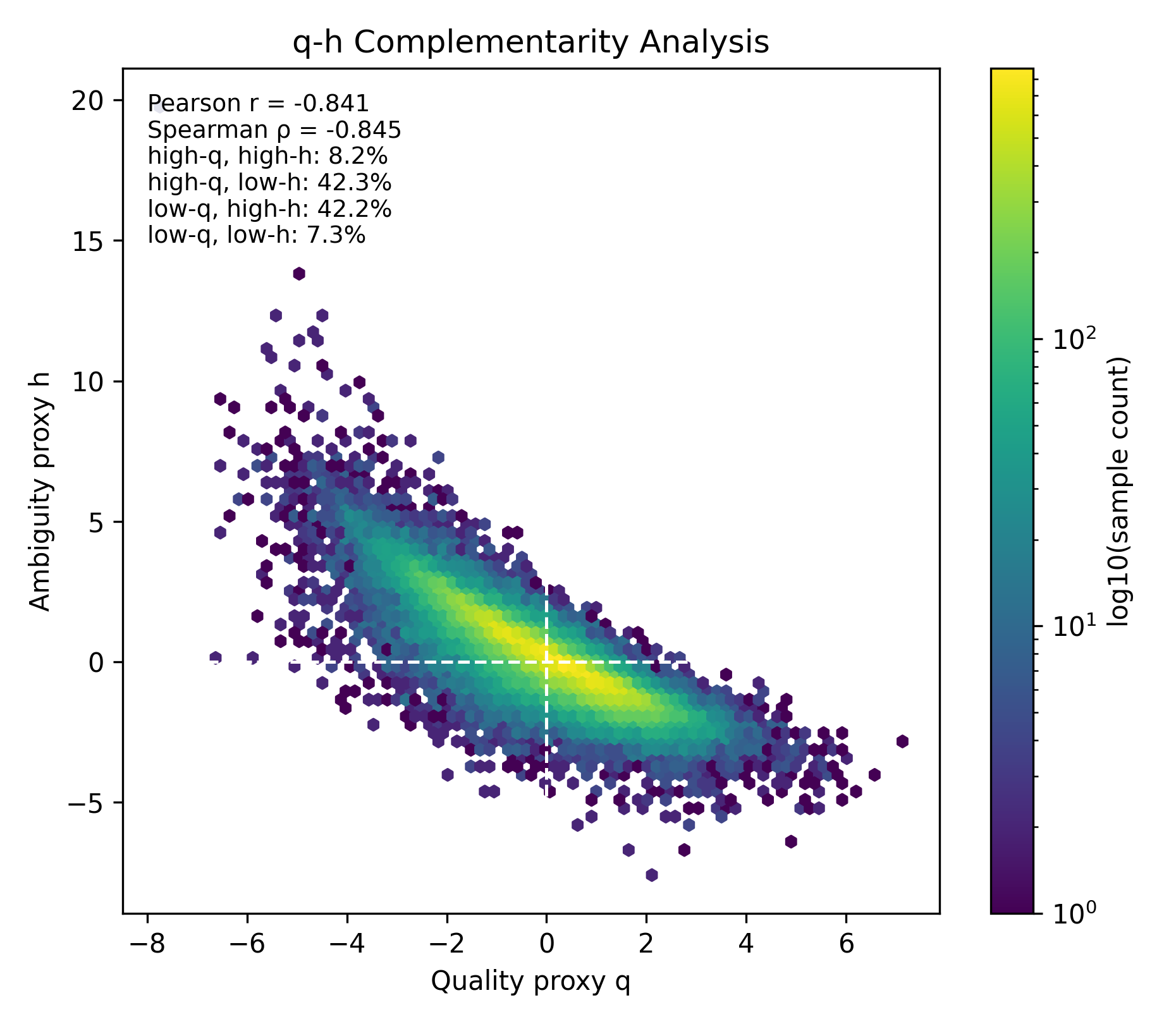}
  \caption{Additional $q$--$h$ analysis. The signal strength proxy $q$ and the ambiguity proxy $h$ are correlated overall, yet the off-diagonal regions (\eg, high-$q$/high-$h$ and low-$q$/low-$h$ samples) show that they are not interchangeable. The quality-adjusted difficulty score therefore combines ambiguity-based ordering with a low-signal penalty, with low-$q$/high-$h$ samples assigned the lowest early-stage priority.}
  \label{fig:qh}
\end{figure}

\noindent \textbf{Complementarity of the Dual Proxies.}
To examine why PICA decouples signal strength $q$ and ambiguity $h$ rather than relying on a single scalar, we visualize their joint distribution in Fig.~\ref{fig:qh}. Although the two proxies are globally correlated, a non-negligible portion of regions violates this trend, occupying the off-diagonal high-$q$/high-$h$ and low-$q$/low-$h$ regimes. These off-diagonal samples show that $q$ and $h$ are correlated but non-substitutable: a region can carry a strong positive signal yet still lie close to a confounding negative, or vice versa. By converting them into the quality-adjusted difficulty score $d$, PICA uses ambiguity for the main easy-to-hard ordering and uses low signal strength as a reliability penalty. This analysis is consistent with the controlled sampler comparison in Table~\ref{tab:single_proxy}.

\noindent \textbf{Rare and Hard Sample Limitation.}
Curriculum sampling raises a practical question: whether average robustness is improved by selecting easier regions more often while rare or hard patterns receive less emphasis. PICA partly avoids this because $\mathcal{L}_{\text{det}}$ and $\mathcal{L}_{\text{ovod}}$ are computed on all valid proposals/regions, and the hard-tier sampling ratio increases with training progress. Still, rare-pattern retention deserves more direct study, especially for low-frequency classes, the hardest $h$ quartile, and failure-prone domains such as Sketch and Tattoo.

\noindent \textbf{Failure Analysis.}
PICA is less effective for some domain shifts. The OV-COCO-O results show clear gains on Painting and Weather, but degradation on Tattoo and little change on Sketch. These domains contain extreme stylization, sparse contours, and texture statistics far from COCO, so the positive signal proxy $q$ can underestimate valid novel regions and the ambiguity proxy $h$ can be dominated by style-induced confounders. This failure mode points to the need for stronger object grounding, style-robust region encoders, or explicit rare-pattern preservation.

\noindent\textbf{Hyperparameter Analysis.}
Table~\ref{tab:hyper} examines the sensitivity of PICA to the initial sampling ratios $(r_E^0,r_M^0,r_H^0)$ and the queue size $M$. For the sampling ratios, the balanced configuration $(r_E^0,r_M^0,r_H^0)=(0.33,0.33,0.33)$ achieves the best performance on both the standard and corrupted evaluation settings. In contrast, configurations strongly biased toward either easy or hard samples
lead to lower performance, indicating that balanced initial coverage of the three difficulty tiers is beneficial.
For the queue size, PICA performs consistently across the evaluated settings, with $M=4096$ achieving the best results. This suggests that a moderate queue size provides a favorable trade-off between sample
diversity and feature freshness.
Additional hyperparameter analyses are provided in the Appendix.

\begin{table}[!t]
  \footnotesize
  \centering
  \caption{Hyperparameter Sensitivity Analysis of PICA on OV-COCO-C. All Results Are Reported on Novel Categories Using the $\text{mAP}^{\text{avg}}_{50}$.}
  \label{tab:hyper}
  \setlength{\tabcolsep}{6pt}
  \renewcommand{\arraystretch}{1.15}
  \begin{tabular}{lc|cc}
    \toprule
    \textbf{Param} & \textbf{Setting} & \textbf{Std.} & \textbf{mAP$^{\text{avg}}_{50}$} \\
    \midrule
    \multirow{5}{*}{$(r_E^0, r_M^0, r_H^0)$}
    & (0.65, 0.30, 0.05) & 36.8 & 20.0 \\
    & (0.60, 0.30, 0.10) & 37.0 & 20.2 \\
    & (0.40, 0.30, 0.30) & 36.9 & 20.1 \\
    & (0.30, 0.30, 0.40) & 36.6 & 19.8 \\
    & \textbf{(0.33, 0.33, 0.33)} & \textbf{37.5} & \textbf{20.7} \\
    \midrule
    \multirow{4}{*}{Queue $M$}
    & 8192 & 36.5 & 20.0 \\
    & 5120 & 36.8 & 20.2 \\
    & 2048 & 36.4 & 19.7 \\
    & \textbf{4096} & \textbf{37.5} & \textbf{20.7} \\
    \bottomrule
  \end{tabular}
\end{table}

\section{Conclusion}
We present PICA for Domain-Generalized Open-Vocabulary Object Detection, suggesting that cross-modal alignment instability is an important bottleneck for OVOD under distribution shifts. Our evaluation protocol and findings reveal a robustness limitation in representative OVOD methods, and indicate that progressive cross-modal alignment regularization can improve novel-category performance across synthetic corruptions and natural OOD settings. However, DG-OVOD remains a challenging problem. This work takes a first step by defining a systematic evaluation benchmark and a feasible training strategy. The current validation focuses on projection-based OVOD with Faster R-CNN-style detectors; broader verification on stronger grounding frameworks, additional real-world robotics/autonomous-driving scenarios, and explicit rare-sample retention metrics remains an important direction for future work.

\bibliographystyle{IEEEtran}
\bibliography{main}

\end{document}